\documentclass[11pt,a4paper]{article}
\usepackage{authblk} 
\usepackage[hyperref]{emnlp-ijcnlp-2019}
\usepackage{times}
\usepackage{latexsym}

\usepackage{graphicx} 
\usepackage{amsfonts} 
\usepackage{mathtools}
\usepackage{booktabs}
\usepackage{subfigure}

\usepackage{url}

\aclfinalcopy 

\title{Alleviating Sequence Information Loss with Data Overlapping and Prime Batch Sizes}

\renewcommand\Affilfont{\small}

\author[$\diamondsuit$]{No{\'e}mien Kocher}
\author[$\spadesuit$]{Christian Scuito}
\author[$\spadesuit$]{Lorenzo Tarantino}
\author[$\heartsuit$]{Alexandros Lazaridis}
\author[$\diamondsuit$,$\clubsuit$]{\\Andreas Fischer}
\author[$\heartsuit$]{Claudiu Musat}
\affil[$\diamondsuit$]{School of Engineering and Architecture of Fribourg, Switzerland, HES-SO, iCoSys Institute}

\makeatletter
\renewcommand\AB@affilsepx{\hspace{.02\linewidth} \protect\Affilfont}
\makeatother

\affil[$\spadesuit$]{Ecole Polytechnique F{\'e}d{\'e}rale de Lausanne (EPFL)}
\affil[$\heartsuit$]{Swisscom}

\makeatletter
\renewcommand\AB@affilsepx{\\
\Affilfont}
\makeatother

\affil[$\clubsuit$]{University of Fribourg}

\makeatletter
\renewcommand\AB@affilsepx{, \protect\Affilfont}
\makeatother

\affil[ ]{\scriptsize \tt \{noemien.kocher|sciutochristian|lorenzotara7\}@gmail.com\protect\\[-1mm]
\scriptsize{\tt\{alexandros.lazaridis|claudiu.musat\}@swisscom.ch, andreas.fischer@hefr.ch}}

\date{}

\begin{document}
\maketitle
\begin{abstract}
In sequence modeling tasks the token order matters, but this information can be partially lost due to the discretization of the sequence into data points. In this paper, we study the imbalance between the way certain token pairs are included in data points and others are not. We denote this a token order imbalance (TOI) and we link the partial sequence information loss to a diminished performance of the system as a whole, both in text and speech processing tasks.
We then provide a mechanism to leverage the full token order information---Alleviated TOI---by iteratively overlapping the token composition of data points. 
For recurrent networks, we use prime numbers for the batch size to avoid redundancies when building batches from overlapped data points.
The proposed method achieved state of the art performance in both text and speech related tasks.
\end{abstract}

\section{Introduction}

Modeling sequences is a necessity. From time series \cite{connor:1994recurrent, lane1999:temporal} to text \cite{sutskever:2011generating} and voice \cite{robinson1994:voivernns,vinyals2012:voicernns2}, ordered sequences account for a large part of the data we process and learn from. The data are discretized and become, in this paradigm, a list of \textit{tokens}.

The key to processing these token sequences is to model the interactions between them. Traditionally \cite{rosenfeld:2000} this has been achieved with statistical methods, like N-grams.

With the advances in computing power and the rebirth of neural networks, the dominant paradigm has become the use of recurrent neural networks (RNNs) \cite{mikolov:2010}. 

The dominance of RNNs has been recently challenged with great success by self-attention based models \cite{vaswani:2017}. Instead of modeling the sequence linearly, Transformer-based models use learned correlations within the input to weight each element of the input sequence based on their relevance for the given task.

\textbf{Series discretization.} 
Both RNNs and self-attention models take as input \textit{data points}---token sequences of a maximum predefined length---and then create outputs for each of them. These tend to be much shorter in size, compared to the size of the full dataset. 
While for humans time seems to pass continuously, this discretization step is important for the machine understanding of the sequence.

\begin{figure}
    \centering
    \includegraphics{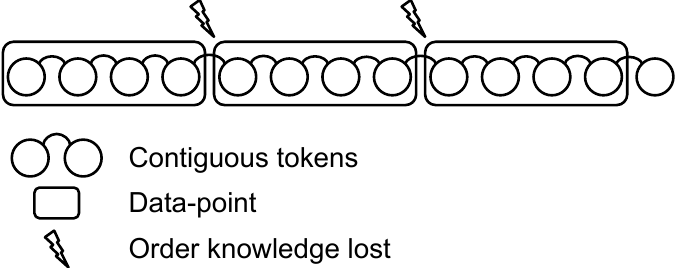}
    \caption{The common way of building data points given a dataset of contiguous tokens. Here we illustrate a dataset with  a contiguous list of 13 tokens, from which we build 3 data points of 4 tokens each. This process keeps the order of the tokens inside the data points, but loses the order information from token pairs that happen to fall between adjacent data points.}
    \label{fig:standard order}
\end{figure}

A side effect of this step is a partial loss of the token order information. As portrayed in Figure~\ref{fig:standard order}, we notice that the token order information within a data point are kept. On the other hand, the knowledge about the token order at the boundaries of data points is lost. We name the situation \textit{Token Order Imbalance (TOI)}. 

As the discretization in Figure~\ref{fig:standard order} is the current standard of sequence processing, we denote this as standard Token Order Imbalance (TOI).
We hypothesize that this loss of information unnecessarily affects the output of the neural network models.

\textbf{Alleviated Token Order Imbalance.}
A first contribution in this work is a mechanism to ensure that all token sequences are taken into account, i.e. every token pair is included in a data point and does not always fall between two data point boundaries. Thus, all sequence information is available for subsequent processing. The proposed method, denoted Alleviated TOI, employs a token offset in the data point creation to create overlapped data point sequences in order to achieve this effect.

\textbf{Batch Creation with Alleviated TOI.}
A second contribution is a strategy for batch creation when using the proposed Alleviated TOI method. We have observed an unintended data redundancy within batches introduced by the overlapped data point sequences. A strategy for avoiding this data redundancy is surprisingly simple but effective: Always use a prime number for the batch size. The intuition behind the prime batch size is that it ensures a good distribution of the batches over the entire dataset. If used naively, the Alleviated TOI policy leads to very similar data points being selected in a batch, which hinders learning. By decoupling the batch size and the token offset used in the token creation, this negative effect is effectively removed.

We then compare the Alleviated TOI with the Standard TOI and show that, on the same dataset and with the same computation allocated, the Alleviated TOI yields better results. The novel TOI reduction method is applicable to a multitude of sequence modeling tasks. We show its benefits in both text and voice processing. We employ several basic and state of the art RNNs as well as Transformers and the results are consistent---the additional information provided by the Alleviated TOI improves the final results in the studied tasks.

For text processing we focus on a well-studied task---language modeling---where capturing the sequence information is crucial. Using Alleviated TOI~(P) with the Maximum Over Softmax (MoS) technique on top of a recurrent cell \cite{Zhilin:17} we get the new state of the art on the Penn-Tree-Bank dataset without fine-tuning with 54.58 perplexity on the test set. We also obtain results comparable to the state of the art on speech emotion recognition on the IEMOCAP \cite{Busso2008IEMOCAPIE} dataset\footnote{To make our results reproducible, all relevant source codes are publicly available at \url{https://github.com/nkcr/overlap-ml}}.

The paper continues with an overview of the related work in Section \ref{sec-related}, a description of the alleviated TOI mechanism in Section \ref{overlapping method} and a detailed description of the batch generation in Section \ref{section: prime}. The experimental design follows in Section \ref{sec-exp} and the results are detailed and interpreted in Section~\ref{sec-results}.

\section{Related work}
\label{sec-related}

At the core of our work is the idea that the way that data samples are provided for training a model can affect speed or capabilities of the model. 
This field is broad and there are several distinct approaches to achieve it. Notable examples include curriculum learning \cite{bengio:09}
and self-paced learning \cite{kumar:10}, where data points for training are selected based on a metric of \textit{easiness} or \textit{hardness}.
In Bayesian approaches \cite{KleinFBHH:16}, the goal is to create sub-samples of data points, whose traits can be extrapolated as the full dataset.

Our work thus differs from the aforementioned methods in the fact that we focus on exploiting valuable but overlooked information from sequences of tokens. 
We change the way data points are generated from token sequences and extend the expressivity of a model by providing an augmented, and well sorted, sequence of data points. This method has a related effect of a randomized-length backpropagation through time (BPTT) \cite{Merity:17}, which yields different data points between epochs. It also resembles classical text data-augmentation methods, such as data-augmentation using thesaurus \cite{zhang:2015text}. 

Our method takes a step forward and proposes a systematic and deterministic approach on building data points that provides the needed variety of data points without the need of randomized-length backpropagation through time (BPTT). This has the effect of producing a text-augmentation without the need of using external resources such as a thesaurus, but only requires the dataset itself. Our method uses a concept of overlapped data points, which can be found in many areas such as data-mining \cite{dong:2007sequence}, DNA sequencing \cite{ng2017:dna2vec}, spectral analysis \cite{ding2000:short}, or temporal data \cite{lane1999:temporal}. In language modeling however, this approach of overlapped data points has not yet been fully exploited.
On the other hand, extracting frame-based acoustic features such as mel-fequency cepstral coefficients (MFCCs) using overlapping windows is a common technique in speech processing and more specifically in automatic speech recognition (ASR) \cite{ChiuSOTA, xiong2016achieving, kim2016power}. We hypothesize that extending the current overlapping technique to a higher level, that is using a sliding overlapping window over the already extracted features, will be proven beneficial.
We believe this to have a positive impact on speech processing tasks such as speech emotion recognition (SER). This is because the emotional load in an spoken utterance expands over larger windows than frame-, phoneme- or syllable-based ones \cite{Frijda1986:emotions}. 

We investigate the proposed method using a simple LSTM model and a small-size Transformer model on the IEMOCAP dataset \cite{Busso2008IEMOCAPIE}, composed of five acted sessions, for a four-class emotions classification and we compare to the state of the art \cite{Mirsamadi2017AutomaticSE} model, a local attention based BiLSTM. \citet{Ramet2018ContextAwareAM} showed in their work a new model that is competitive to the one previously cited, following a cross-valiadation evaluation schema. For a fair comparison, in this paper we focus on a non-cross-valiation schema and thus compare our results to the work of \citet{Mirsamadi2017AutomaticSE}, where a similar schema is followed using as evaluation set the fifth session of IEMOCAP database. It is noteworthy that with a much simpler method than presented in  \citet{Ramet2018ContextAwareAM}, we achieve comparable results, underscoring the importance of the proposed method for this task as well.

\section{Alleviated Token Order Imbalance} 
\label{overlapping method}

Let a token pair denote an ordered pair of tokens---for instance token A followed by token B, as in the sequence $"A B C D E F G ..."$.
When splitting a token sequence into data points $"D1, D2, .."$, if the split is fixed, as in $D1$ always being equal to $"A B C"$, $D2$ always being equal to $"D E F"$, etc., then the information contained in the order of tokens $C$ and $D$ for instance is partially lost. This occurs as there is no data point that contains this token pair explicitly. We call the $"C D"$ token pair a \textit{split token pair} and its tokens, $C$ and $D$, are denoted as \textit{split tokens}.

In its most extreme form, split token pair order information is lost completely. In other cases, it is partially taken into account implicitly. In recurrent cells, for instance, the internal state of the cell allows for the order information of split tokens pairs to be used. This is due to the serial processing of the data points containing the split tokens.

As some token pairs are taken into account fully, others partially and others not at all, we denote this situation as \textit{token order imbalance} (TOI).

In this paper, we propose to alleviate the TOI by means of overlapping sequences of data points. The aim is to avoid the loss of information between the last token of a data point and the first token of its subsequent data point. Instead of splitting the sequence of tokens only once, we repeat this process multiple times using different offsets. Each time we subdivide the sequence of tokens with a new offset, we include the links that were missing in the previous step. Finally, the overlapping sequences of data points are concatenated into a single sequence, forming the final dataset.

\begin{figure}
    \centering
    \includegraphics{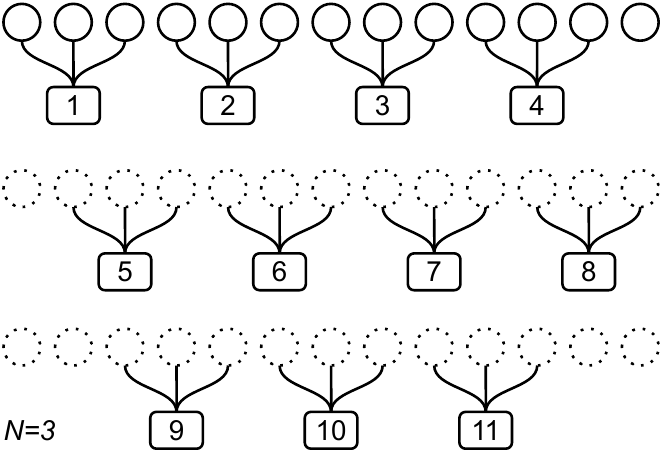}
    \caption{Illustration of an Alleviated TOI~(3) made from a single contiguous list of 13 tokens. With a Standard TOI and $N$=3 (ie. 3 tokens per data point), a contiguous list of 13 tokens would produce 4 data points, which is illustrated by the first overlapped sequence. Here, an Alleviated TOI~(3) splits the contiguous list of tokens 3 times with each time a different offset ($0,1,2$ respectively). This finally leads to a list of 11 data points coming from the 3 appended overlapped sequences.}
    \label{fig:overlapped sequences}
\end{figure}

Figure~\ref{fig:overlapped sequences} illustrates an Alleviated TOI~(3), which means the sequence of data points is split three times instead of only once, producing 3 overlapped sequences that will then be concatenated.

Our Alleviated TOI~(P) method is detailed in the pseudo-code below, where {\small\verb|olp_sequence|} holds an overlapped sequence and $P$ is the number of times we subdivide the sequence of tokens with a different offset:

\vspace{1em}
\begin{small}
\begin{verbatim}
Let N = Number of tokens per data point
    P = Number of overlapped sequences
    Step = N / P
DataPoints = empty list
FOR i = 0..P-1
  olp_sequence = create data points 
                 from Dataset with 
                 offset (i * Step)
  Add olp_sequence to DataPoints
RETURN DataPoints
\end{verbatim}
\end{small}

When we apply an Alleviated TOI~(P), this means that we are going to create $P$ times a sequence of data points with different offsets. Therefore, the final dataset will be the concatenation of $P$ repetitions of the original dataset, with data points shifted by a specific and increasing offset at token level for each repetition.

For example, given a sequence $S_1$ with $N = 70$ tokens per data point and an Alleviated TOI~(P) with $P = 10$, the step size will be $\frac{N}{P}=7$ tokens. Therefore, starting from the sequence $S_1$, nine additional sequences of data points will be created: $S_2$ starting from token 7, $S_3$ starting from token 14, $S_4$ starting from token 21 and so on until $S_{10}$.

When using Alleviated TOI~(P), with $P$ smaller than the data point size, within an epoch, a split token pair---that is a token pair that is split in the original data point splitting---becomes part of a data point $P-1$ times. A token pair that is never split will be part of the data point $P$ times.

We can thus define a \textit{token order imbalance ratio} that describes the imbalance between the number of times we include split token pairs and the number of times we include pairs that are not split: 
\[
(P-1) / P
\]

We notice that the higher $P$, the closer the ratio becomes to 1. We hypothesize that the closer the ratio becomes to 1, the better we leverage the information in the dataset. We thus expect that for higher values of P the Alleviated TOI~(P) method will outperform versions with lower values, with Alleviated TOI~(1) being the Standard TOI, which is now prevalent.

We quantify the additional computational cost of Alleviated TOI~(P). Since our method only results in $P$ (shifted) repetitions of the dataset, each epoch using the augmented dataset would take $\sim P$ times longer than an epoch over the original dataset. Therefore, we ensure fair comparison by allowing baseline models to train for $P$ times more epochs than a model using Alleviated TOI~(P).

\section{Batch Creation with Alleviated TOI}
\label{section: prime}

Series discretization may also occur at higher levels than data points, in particular when building batches for mini-batch training of neural networks. We can distinguish two types of batches, i.e. \textit{sequential} and \textit{distributed} batches. The former keep the data point sequences intact, thus creating split token pairs only between two consecutive batches. The latter distribute data points from different parts of the dataset to approximate the global distribution, thus creating split token pairs between all data points in batches.

In principle, our proposed method alleviates the TOI in both cases, since multiple overlapping sequences of data points are generated. However, we have observed an unintended interference with the batch creation in the case of distributed batches. In this section we explain the problem in detail and propose a simple but effective solution---choosing a prime batch size.

\begin{figure}
    \centering
    \includegraphics{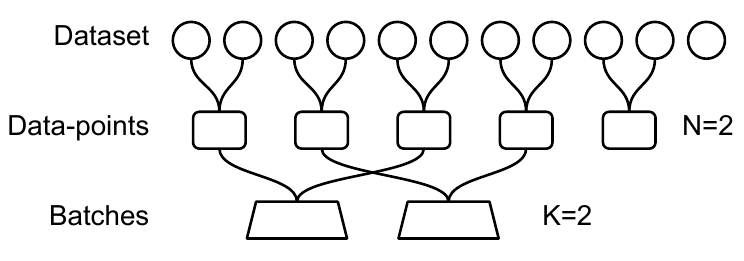}
    \caption{Three levels of data representation used to create distributed batches. The dataset is a sequence of tokens on which data points are built by splitting the sequence into subsequences of N tokens. Batches of K data points are then built by subdividing the sequence of data points into K equal parts. Here, the first part contains the first two data points, the second part the following two, and the last data point is dropped. Each batch then uses one element of each part.}
    \label{fig:three levels}
\end{figure}

Figure~\ref{fig:three levels} illustrates the three levels of data representation in the case of distributed batches. Data points are built from $N$ consecutive tokens to capture the sequential information. Batches are then built from $K$ parts of the data point sequence to capture the global distribution. An example of this approach is the batching procedure used in ~\citet{zoph2016:neural,Merity:17,Zhilin:17,zolna2017:fraternal} for word language modeling, where the basic token is a word.

The batching mechanism can be seen as building a 2-dimensional matrix, where each row contains a batch. Consider a sequence of $M$ data points and a batch size of $K$. In order to build batches, the data points are split into $K$ parts, represented as $\frac{M}{K} \times 1$ column vectors. They are concatenated to form a $\frac{M}{K} \times K$ matrix, such that the rows correspond to batches.

\begin{figure*}
  \centering
  \mbox{
    \subfigure[Matrix of batches with batch size of 20
    \label{fig:non prime matrix of pixel}]{
        \includegraphics[]{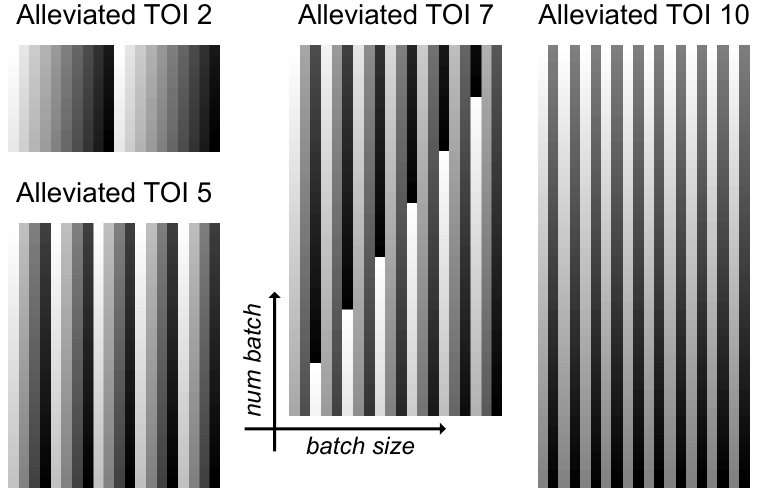}
    }\quad
    \subfigure[Matrix of batches with batch size of 19
    \label{fig: prime matrix of pixel}]{
        \includegraphics[]{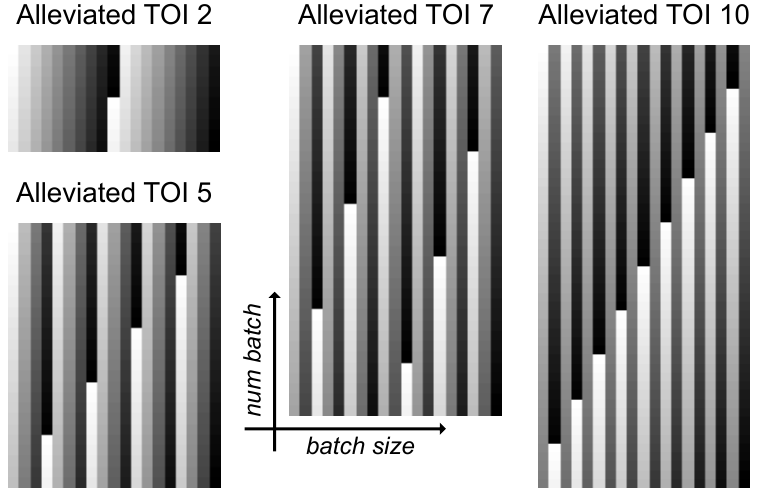}
    }
  }
  \caption{Illustrations of the 2D matrix of batches with different $P$-values of Alleviated TOI (P). On the left we used a batch size of 20 and on the right we used a prime batch size of 19. Each data point is a pixel and each row is a batch. The grayscale value models the proximity of the data points with respect to the dataset. Therefore, two pixels with similar color represents two data points that are close in the dataset. The illustrations demonstrate how different values of $P$ affect the content of the batches, which can lack a good distribution over the dataset. Ideally, each row should contain a gradient of different grayscale values.
  We can observe how using a prime batch size affects the distribution of data points within the batches, where the matrices on the right offer a better distribution. This effect is especially well visible for the Alleviated TOI 10.
  }
  \label{fig:datapoints as pixels}
\end{figure*}

\begin{figure}
    \centering
    \includegraphics[]{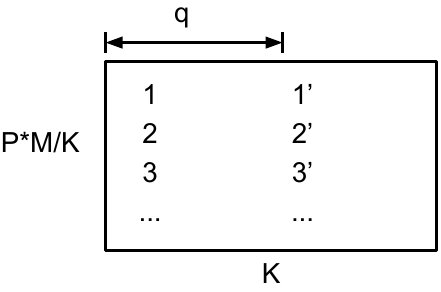}
    \caption{Data point repetition with period $q$ for $M$ data points, $K$ batches, and Alleviated TOI~(P). Data point 1' is the same as data point 1 with a token offset.}
    \label{fig:period}
\end{figure}

\begin{table}[t]
\centering
\begin{tabular}{@{}ccc@{}}
\toprule
\textbf{$P$} & \textbf{Period $q$} & \textbf{Repetitions} \\ \midrule
2                    & 10             & 2              \\
5                    & 4            & 5               \\
7                    & 20            & 1              \\
10                   & 2             & 10               \\ \bottomrule
\end{tabular}
\caption{Data point repetition with period $q$ for batch size $K=20$ and Alleviated~TOI (P).}
\label{finite period}
\end{table}

When applying the proposed Alleviated TOI~(P) method (see Section~\ref{overlapping method}), we augment the original dataset to a total of $P \cdot M$ data points, adding additional data points with token offsets. Therefore, the $\frac{P \cdot M}{K} \times K$ matrix used for batch creation may contain repeated data points within the same batch as illustrated in Figure~\ref{fig:period}. A \textit{repeated data point} differs from the previous data point only marginally due to the token offset. This redundancy can be problematic, as the batches are not well-distributed over the entire dataset anymore.

With respect to the batch matrix, a repeated data point occurs iff $\frac{P \cdot M}{K} \cdot q = n \cdot M$ with \textit{period} $q < K$ and $q,n \in \mathbb{N}$. This is equivalent to
\[\frac{P}{K} \cdot q = n,\ \ \ q < K, q,n \in \mathbb{N}\]
independent of the number of data points $M$. A repetition thus occurs iff the greatest common divisor (GCD) of $P$ and $K$ is larger than 1. Otherwise, for GCD$(P,K) = 1$ a data point repeats only after period $q=K$, i.e. there is no repetition within the same batch.

Table~\ref{finite period} lists exemplary periods for a batch size of $K=20$ and different values of $P$ for the Alleviated TOI (P). The worst case is $P=10$ with $10$ repetitions of the same data point within the same batch and the best case is $P=7$, which avoids any redundancy because the GCD of $P$ and $K$ is 1. Figure~\ref{fig:datapoints as pixels} illustrates the repetition with grayscale values, where similar grayscale values indicate that two data points are close within the original data points sequence.

In general, while we aim for large values of $P$ for reducing the TOI, a simple solution for avoiding redundancy within batches is to choose a prime number for the batch size $K$.

\section{Experimental Setup}
\label{sec-exp}
To validate the generalization capability of the proposed technique, we apply it on both text and speech related tasks. We thus run the Alleviated TOI~(P) with language modeling (text) and emotion recognition (speech). The text datasets used are Penn-Tree-Bank (PTB) \cite{Marcus:1993} as preprocessed in \citet{Mikolov:11}, Wikitext-2 (WT2), and Wikitext-103 (WT103) \cite{Merity:16}.
The speech dataset is the IEMOCAP database \cite{Busso2008IEMOCAPIE}, a collection of more than 12 hours of recorded emotional speech of 10 native-English speakers, men and women. The audio data is filtered down to 5.5 hours containing only \textit{angry}, \textit{happy}, \textit{neutral} and \textit{sad} utterances.

\subsection{TOI in Language Modelling}

For language modeling, we use three different methods:
\begin{itemize}
    \item A simple LSTM that does not benefit from extensive hyper-parameter optimization.
    \item An Average Stochastic Gradient Descent Weight-Dropped LSTM (AWD-LSTM) as described in \citet{Merity:17}, with the same hyper-parameters.
    \item The latest State-of-the-Art model: Mixture of Softmaxes (MoS) \cite{Zhilin:17}.
\end{itemize} 
We compare our results against the original process of building data points, i.e. Standard TOI, and use the same computation load allocated for each experiment. We use the same set of hyper-parameters as described in the base papers, except for the batch size with Alleviated TOI~(P), where we use a prime batch size in order to prevent any repetitions in batches, as described in Section \ref{section: prime}. 
That is, on the PTB dataset, we use a sequence length of 70 for all the models. For the Simple LSTM and AWD-LSTM, we use a batch size of 20 and a hidden size of 400.
AWD-LSTM and MoS are trained on 1000 epochs, and the Simple LSTM on 100 epochs. For the MoS model, embedding size used is 280, batch size 12, and hidden size 980.
All the models use SGD as the optimizer.

We set up experiments to compare 4 different token order imbalance setups: Extreme TOI, Inter-batch TOI, Standard TOI, and Alleviated TOI~(P).

\paragraph{Extreme TOI} The Extreme TOI setup builds batches using a random sequence of data points. This removes any order inside the batches (i.e. among data points within a batch), and among batches. 

\paragraph{Inter-batch TOI} In the Inter-batch TOI setup, batches are built using an ordered sequence of data points, but the sequence of batches is shuffled. This keeps the order inside batches, but removes it among batches. Looking at the 2D matrix of batches, in Figure~\ref{fig:datapoints as pixels}, this results in shuffling the rows of the matrix.

\paragraph{Standard TOI} In the Standard TOI setup, the process of building batches is untouched, as described in section \ref{overlapping method}. This keeps the order inside, and among batches. 

\paragraph{Alleviated TOI~(P)} In the Alleviated TOI~(P) setup, we apply our proposed TOI reduction by creating $P$ overlapped data point sequences (see Sections~\ref{overlapping method} and~\ref{section: prime}). This strategy not only keeps the order inside and among batches, but it also restores the full token order information in the dataset.

\subsection{TOI in Speech Emotion Recognition}

For Speech Emotion Recognition (SER) we use two different models: the encoder of the Transformer \cite{vaswani:2017} followed by convolutional layers, and the simple LSTM used in text domain case. Since the Transformer is stateless and uses self-attention instead, we are able to investigate the effect of Alleviated TOI~(P) independently of LSTM cells.

As with language modeling, we set up experiments to compare the 4 different token order imbalance strategies: Extreme TOI, Inter-batch TOI, Standard TOI, and Alleviated TOI~(P). 

We apply the methodology used in text on the SER task, using the simple LSTM and a window size of 300 frames. In this case, a data point, instead of being a sequence of words, is a sequence of frames coming from the same utterance. Each frame is described by a 384-dimensional features vector. OpenSMILE \cite{Eyben:2013:RDO:2502081.2502224} is used for extracting the features. We opt for the IS09 features set \cite{Schuller2009Interspeech} as proposed by \citet{Ramet2018ContextAwareAM} and commonly used for SER.

Finally, to investigate the effect of the Alleviated TOI~(P) strategy independently of LSTM cells, we design a final experiment in the SER task. We investigate whether or not we have improved results as we increase $P$, the number of overlapped data point sequences in a stateless scenario. For this reason, we use the Transformer model described above.

\section{Experimental Results}
\label{sec-results}
\subsection{Language Modelling}

Table~\ref{table: orders with awd model} compares the 4 token order imbalance strategies using the AWD model and three text datasets. We use the test perplexity after the same equivalent number of epochs. The different Alleviated TOI (P) experiments use a different number of overlapped sequence: An Alleviated TOI (P) means building and concatenating $P$ overlapped sequences. Our results indicate that an Alleviated TOI (P)
is better than the Standard TOI, which is better than an Extreme or Inter-batch TOI. We note a tendency that higher values of $P$ lead to better results, which is in accordance with our hypothesis that a higher TOI ratio $(P-1)/P$ improves the results.

\begin{table}[t]
\centering

\begin{tabular}{@{}llll@{}}
\toprule
  Experiment & PTB & WT2 & WT103\\
  \midrule
  Extreme TOI       & 63.49          & 73.52          & 36.19 \\
  Inter-batch TOI   & 64.20          & 72.61          & 36.39 \\
  Standard TOI      & \textbf{58.94}         & \textbf{65.86} & \textbf{32.94} \\
  Alleviated TOI 2  & 57.97          & 65.14          & 32.98 \\
  Alleviated TOI 5  & 57.14        & 65.11          & 33.07 \\
  Alleviated TOI 7  & 57.16         & 64.79          & 32.89 \\
  Alleviated TOI 10 & \textbf{56.46} & \textbf{64.73} & \textbf{32.85} \\
\bottomrule
\end{tabular}
\caption{
    \label{table: orders with awd model}
    Perplexity score (PPL) comparison of the AWD model, on the three datasets, with batch sizes $K=20$~(PTB), $K=80$~(WT2) and $K=60$~(WT103), with different levels of Token Order Imbalance (TOI). With Alleviated TOI~(P), we use a prime batch size of $K=19$~(PTB), $K=79$~(WT2) and $K=59$~(WT103).
  }
\end{table}

\paragraph{Comparison with State of the Art and Simple LSTM.}

With the MoS model and an Alleviated TOI, we improve the current state of the art without fine tuning for the PTB dataset with 54.58 perplexity on the test set. Table~\ref{table: sota with mos} demonstrates how models can be improved by applying our Alleviated TOI method on 2 latest state-of-the-art models: AWD-LSTM \cite{Merity:17} and AWD-LSTM-MoS \cite{Zhilin:17}, and the Simple LSTM model. We compare the results with the same hyper-parameters used on the original papers with the only exception of the batch size, that must be prime. To ensure fairness, we allocate the same computational resources for the base model as well the model with Alleviated TOI, i.e. we train with the equivalent number of epochs.

\begin{table}[t]
\centering
\begin{tabular}{@{}llll@{}}
\toprule
  Model & test ppl\\
  \midrule
  AWD-LSTM \cite{Merity:17} & 58.8 \\
  AWD-LSTM + Alleviated TOI & 56.46 \\
  AWD-LSTM-MoS \cite{Zhilin:17} & 55.97 \\
  AWD-LSTM-MoS + Alleviated TOI & 54.58 \\
  Simple-LSTM & 75.36 \\
  Simple-LSTM + Alleviated TOI & 74.44 \\
\bottomrule
\end{tabular}
\caption{
    \label{table: sota with mos}
    Comparison between state-of-the-art models \cite{Merity:17, Zhilin:17} and a Simple LSTM, and the same models with Alleviated TOI. The comparison highlights how the addition of Alleviated TOI is able to improve state-of-the-art models, as well as a simple model that does not benefit from extensive hyper-parameter optimization.
  }
\end{table}

\paragraph{Comparison without prime batch size.}

In Table~\ref{table: prime comparison} we demonstrate how using a prime batch size with Alleviated TOI (P) actually impacts the scores. We compare the scores of a prime batch size $K=19$ with the scores of the original batch size $K=20$ for the AWD model with Alleviated TOI (P). When using a prime batch size, we observe consistent and increasing results as $P$ increases. This is due to the good distribution of data points in the batches regardless of the value of $P$, which is visible in Figure~\ref{fig: prime matrix of pixel} where each row contains a high diversity of grayscale values. With the original batch size $K=20$, we observe a strong performance for $P=7$, but a low performance for $P=10$. Again, this effect is related to the distribution of data points in the batches, which is visible in Figure~\ref{fig:non prime matrix of pixel}. The matrix with $P=7$ shows a good distribution---corresponding to the strong performance---and the matrix with $P=10$ shows that each row contains a low diversity of data points.

\begin{table}[t]
\centering
\begin{tabular}{@{}llll@{}}
\toprule
  Experiment & K=20 & K=19\\
  \midrule
  Alleviated TOI 2  & 59.37 & 57.97 \\
  Alleviated TOI 5  & 60.50 & 57.14 \\
  Alleviated TOI 7  & \textbf{56.70} & 57.16 \\
  Alleviated TOI 10 & 65.88 & \textbf{56.46} \\
\bottomrule
\end{tabular}
\caption{
    \label{table: prime comparison}
    Perplexity score (PPL) comparison on the PTB dataset and the AWD model. We use two different values for the batch size $K$ --- the original one with $K=20$, and a prime one with $K=19$. The results directly corroborate the observation portrayed in Figure~\ref{fig:datapoints as pixels}, where the obtained score is related to the diversity of grayscale values in each row.
  }
\end{table}


\subsection{Speech Emotion Recognition Results}

The results on the IEMOCAP database are evaluated in terms of weighted (WA) and unweighted accuracy (UA). The first metric is the accuracy on the entire evaluation dataset, while the second is the average of the accuracies of each class of the evaluation set. UA is often used when the database is unbalanced, which is true in our case, since the \textit{happy} class has a total duration that is half of the second smallest class in speech duration.

Table~\ref{table: lstm on iemocap} shows that our proposed method brings value in the speech related task as well. When choosing the Extreme TOI instead of the Standard TOI approach we observe a smaller effect than in text related task: this is due to the different nature of the text datasets (large "continuous" corpuses) and the IEMOCAP one (composed of shorter utterances). The fact that we can still observe improvements on a dataset with short utterances is a proof of the robustness of the method.

\begin{table}[t]
\centering

\begin{tabular}{@{}lll@{}}
\toprule
  Experiment & WA & UA \\
  \midrule
  Extreme TOI (15k steps) & 0.475 & 0.377 \\
  Inter-batch TOI (15k steps) & 0.478 & 0.386 \\
  Standard TOI (15k steps) & 0.486 & 0.404 \\
  Alleviated TOI (15k steps) & \textbf{0.553} & \textbf{0.489} \\
  Alleviated TOI (60 epochs) & \textbf{0.591} & \textbf{0.523} \\
\bottomrule
\end{tabular}
\caption{
    \label{table: lstm on iemocap}
    Token order imbalance (TOI) comparison for the IEMOCAP dataset on a SER task using \textit{angry, happy, neutral} and \textit{sad} classes with a simple LSTM model. 
  }
\end{table}

A greater effect is obtained when we increase the size of the dataset with the proposed Alleviated TOI~(P) approach: Due to the increasing offset at each overlapped sequence, the data fed into the model contains utterances where the emotions are expressed in slightly different ways. For this reason, the performance notably increases.

Table~\ref{table: transformer overlapping on iemocap - epochs} reports the result of a final experiment that aims to investigate the effect of Alleviated TOI~(P) independently of LSTM cells. For each Alleviated TOI (P) setup and Standard TOI described in Table~\ref{table: transformer overlapping on iemocap - epochs}, we repeat the training and evaluation for each of the following window sizes: 100, 200, 300, 400 and 500 frames. The previously described Transformer model is used in these experiments.
The results reported in Table~\ref{table: transformer overlapping on iemocap - epochs} are the mean $\pm$ the standard deviation computed for different P-values of Alleviated TOI~(P).

\begin{table}[t]
\centering
\resizebox{\linewidth}{!}{%
\begin{tabular}{@{}lll@{}}
\toprule
  Experiment & WA (60 epochs) & UA (60 epochs)\\
  \midrule
  Alleviated TOI 1 & 0.591$\pm$0.012 & 0.543$\pm$0.021 \\
  Alleviated TOI 2 & 0.594$\pm$0.007 & 0.549$\pm$0.016 \\
  Alleviated TOI 3 & 0.605$\pm$0.018 & 0.563$\pm$0.024 \\
  Alleviated TOI 5 & 0.608$\pm$0.015 & 0.562$\pm$0.028 \\
  Alleviated TOI 10 & \textbf{0.617$\pm$0.015} & \textbf{0.571$\pm$0.024} \\
  \midrule
  Local attention & \textbf{0.635} & \textbf{0.588} \\
\bottomrule
\end{tabular}}
\caption{
    \label{table: transformer overlapping on iemocap - epochs}
    Token order imbalance (TOI) comparison for the IEMOCAP dataset on a SER task using \textit{angry, happy, neutral and sad} classes for 60 epochs using the Transformer model.
  }
\end{table}

The last line of Table~\ref{table: transformer overlapping on iemocap - epochs} refers to \citet{Mirsamadi2017AutomaticSE} results. We want to highlight the fact that the goal of these experiments is to show the direct contribution of the Alleviated TOI technique for a different model. For this reason we use a smaller version of the Transformer in order to reduce the computational cost. We believe that with a more expressive model and more repetitions, the proposed method may further improve the results.

The results from Table~\ref{table: transformer overlapping on iemocap - epochs} demonstrate 
that, as we increase the value of $P$, more significant improvements are achieved. This is in accordance with our hypothesis that a higher TOI ratio $(P-1)/P$ improves the results.

\section{Conclusions}
In this work, the importance of overlapping and token order in sequence modelling tasks were investigated.
Series discretization is an essential step in machine learning processes which nonetheless can be responsible for the loss of the continuation of the tokens, through the token order imbalance (TOI) phenomenon.
The proposed method, Alleviated TOI, has managed to overcome this drawback and ensures that all token sequences are taken into account.
The proposed method was validated in sequence modelling tasks both in the text and speech domain outperforming the state of the art techniques.

\bibliography{emnlp-ijcnlp-2019}

\begin{thebibliography}{31}
\expandafter\ifx\csname natexlab\endcsname\relax\def\natexlab#1{#1}\fi

\bibitem[{Bengio et~al.(2009)Bengio, Louradour, Collobert, and
  Weston}]{bengio:09}
Yoshua Bengio, J{\'e}r{\^o}me Louradour, Ronan Collobert, and Jason Weston.
  2009.
\newblock Curriculum learning.
\newblock In \emph{Proceedings of the 26th annual international conference on
  machine learning}, pages 41--48. ACM.

\bibitem[{Busso et~al.(2008)Busso, Bulut, Lee, Kazemzadeh, Provost, Kim, Chang,
  Lee, and Narayanan}]{Busso2008IEMOCAPIE}
Carlos Busso, Murtaza Bulut, Chi-Chun Lee, Abe Kazemzadeh, Emily~Mower Provost,
  Samuel Kim, Jeannette~N. Chang, Sungbok Lee, and Shrikanth Narayanan. 2008.
\newblock Iemocap: interactive emotional dyadic motion capture database.
\newblock \emph{Language Resources and Evaluation}, 42:335--359.

\bibitem[{Chiu et~al.(2018)Chiu, Sainath, Wu, Prabhavalkar, Nguyen, Chen,
  Kannan, Weiss, Rao, Gonina et~al.}]{ChiuSOTA}
Chung-Cheng Chiu, Tara~N Sainath, Yonghui Wu, Rohit Prabhavalkar, Patrick
  Nguyen, Zhifeng Chen, Anjuli Kannan, Ron~J Weiss, Kanishka Rao, Ekaterina
  Gonina, et~al. 2018.
\newblock State-of-the-art speech recognition with sequence-to-sequence models.
\newblock In \emph{2018 IEEE International Conference on Acoustics, Speech and
  Signal Processing (ICASSP)}, pages 4774--4778. IEEE.

\bibitem[{Connor et~al.(1994)Connor, Martin, and Atlas}]{connor:1994recurrent}
Jerome~T Connor, R~Douglas Martin, and Les~E Atlas. 1994.
\newblock Recurrent neural networks and robust time series prediction.
\newblock \emph{IEEE transactions on neural networks}, 5(2):240--254.

\bibitem[{Ding et~al.(2000)Ding, Bressler, Yang, and Liang}]{ding2000:short}
Mingzhou Ding, Steven~L Bressler, Weiming Yang, and Hualou Liang. 2000.
\newblock Short-window spectral analysis of cortical event-related potentials
  by adaptive multivariate autoregressive modeling: data preprocessing, model
  validation, and variability assessment.
\newblock \emph{Biological cybernetics}, 83(1):35--45.

\bibitem[{Dong and Pei(2007)}]{dong:2007sequence}
Guozhu Dong and Jian Pei. 2007.
\newblock \emph{Sequence data mining}, volume~33.
\newblock Springer Science \& Business Media.

\bibitem[{Eyben et~al.(2013)Eyben, Weninger, Gross, and
  Schuller}]{Eyben:2013:RDO:2502081.2502224}
Florian Eyben, Felix Weninger, Florian Gross, and Bj\"{o}rn Schuller. 2013.
\newblock \href {https://doi.org/10.1145/2502081.2502224} {Recent developments
  in opensmile, the munich open-source multimedia feature extractor}.
\newblock In \emph{Proceedings of the 21st ACM International Conference on
  Multimedia}, MM '13, pages 835--838, New York, NY, USA. ACM.

\bibitem[{Frijda(1986)}]{Frijda1986:emotions}
Nico~H. Frijda. 1986.
\newblock \emph{The Emotions}.
\newblock Cambridge University Press.

\bibitem[{Kim and Stern(2016)}]{kim2016power}
Chanwoo Kim and Richard~M Stern. 2016.
\newblock Power-normalized cepstral coefficients (pncc) for robust speech
  recognition.
\newblock \emph{IEEE/ACM Transactions on Audio, Speech and Language Processing
  (TASLP)}, 24(7):1315--1329.

\bibitem[{Klein et~al.(2016)Klein, Falkner, Bartels, Hennig, and
  Hutter}]{KleinFBHH:16}
Aaron Klein, Stefan Falkner, Simon Bartels, Philipp Hennig, and Frank Hutter.
  2016.
\newblock \href {http://arxiv.org/abs/1605.07079} {Fast bayesian optimization
  of machine learning hyperparameters on large datasets}.
\newblock \emph{CoRR}, abs/1605.07079.

\bibitem[{Kumar et~al.(2010)Kumar, Packer, and Koller}]{kumar:10}
M~Pawan Kumar, Benjamin Packer, and Daphne Koller. 2010.
\newblock Self-paced learning for latent variable models.
\newblock In \emph{Advances in Neural Information Processing Systems}, pages
  1189--1197.

\bibitem[{Lane and Brodley(1999)}]{lane1999:temporal}
Terran Lane and Carla~E Brodley. 1999.
\newblock Temporal sequence learning and data reduction for anomaly detection.
\newblock \emph{ACM Transactions on Information and System Security (TISSEC)},
  2(3):295--331.

\bibitem[{Marcus et~al.(1993)Marcus, Marcinkiewicz, and
  Santorini}]{Marcus:1993}
Mitchell~P. Marcus, Mary~Ann Marcinkiewicz, and Beatrice Santorini. 1993.
\newblock \href {http://dl.acm.org/citation.cfm?id=972470.972475} {Building a
  large annotated corpus of english: The penn treebank}.
\newblock \emph{Comput. Linguist.}, 19(2):313--330.

\bibitem[{Merity et~al.(2017)Merity, Keskar, and Socher}]{Merity:17}
Stephen Merity, Nitish~Shirish Keskar, and Richard Socher. 2017.
\newblock \href {http://arxiv.org/abs/1708.02182} {Regularizing and optimizing
  {LSTM} language models}.
\newblock \emph{CoRR}, abs/1708.02182.

\bibitem[{Merity et~al.(2016)Merity, Xiong, Bradbury, and Socher}]{Merity:16}
Stephen Merity, Caiming Xiong, James Bradbury, and Richard Socher. 2016.
\newblock \href {http://arxiv.org/abs/1609.07843} {Pointer sentinel mixture
  models}.
\newblock \emph{CoRR}, abs/1609.07843.

\bibitem[{Mikolov et~al.(2011)Mikolov, Deoras, Kombrink, Burget, and
  {\v{C}}ernock{\`y}}]{Mikolov:11}
Tom{\'a}{\v{s}} Mikolov, Anoop Deoras, Stefan Kombrink, Luk{\'a}{\v{s}} Burget,
  and Jan {\v{C}}ernock{\`y}. 2011.
\newblock Empirical evaluation and combination of advanced language modeling
  techniques.
\newblock In \emph{Twelfth Annual Conference of the International Speech
  Communication Association}.

\bibitem[{Mikolov et~al.(2010)Mikolov, Karafi{\'a}t, Burget,
  {\v{C}}ernock{\`y}, and Khudanpur}]{mikolov:2010}
Tom{\'a}{\v{s}} Mikolov, Martin Karafi{\'a}t, Luk{\'a}{\v{s}} Burget, Jan
  {\v{C}}ernock{\`y}, and Sanjeev Khudanpur. 2010.
\newblock Recurrent neural network based language model.
\newblock In \emph{Eleventh annual conference of the international speech
  communication association}.

\bibitem[{Mirsamadi et~al.(2017)Mirsamadi, Barsoum, and
  Zhang}]{Mirsamadi2017AutomaticSE}
Seyedmahdad Mirsamadi, Emad Barsoum, and Cha Zhang. 2017.
\newblock Automatic speech emotion recognition using recurrent neural networks
  with local attention.
\newblock \emph{2017 IEEE International Conference on Acoustics, Speech and
  Signal Processing (ICASSP)}, pages 2227--2231.

\bibitem[{Ng(2017)}]{ng2017:dna2vec}
Patrick Ng. 2017.
\newblock dna2vec: Consistent vector representations of variable-length k-mers.
\newblock \emph{arXiv preprint arXiv:1701.06279}.

\bibitem[{Ramet et~al.(2018)Ramet, Garner, Baeriswyl, and
  Lazaridis}]{Ramet2018ContextAwareAM}
Gaetan Ramet, Philip~N. Garner, Michael Baeriswyl, and Alexandros Lazaridis.
  2018.
\newblock Context-aware attention mechanism for speech emotion recognition.
\newblock \emph{2018 IEEE Spoken Language Technology Workshop (SLT)}, pages
  126--131.

\bibitem[{Robinson(1994)}]{robinson1994:voivernns}
Anthony~J. Robinson. 1994.
\newblock \href {https://doi.org/10.1109/72.279192} {An application of
  recurrent nets to phone probability estimation}.
\newblock \emph{Trans. Neur. Netw.}, 5(2):298--305.

\bibitem[{Rosenfeld(2000)}]{rosenfeld:2000}
Ronald Rosenfeld. 2000.
\newblock Two decades of statistical language modeling: Where do we go from
  here?
\newblock \emph{Proceedings of the IEEE}, 88(8):1270--1278.

\bibitem[{Schuller et~al.(2009)Schuller, Steidl, and
  Batliner}]{Schuller2009Interspeech}
Bj{\"o}rn Schuller, Stefan Steidl, and Anton Batliner. 2009.
\newblock The interspeech 2009 emotion challenge.
\newblock In \emph{Tenth Annual Conference of the International Speech
  Communication Association}.

\bibitem[{Sutskever et~al.(2011)Sutskever, Martens, and
  Hinton}]{sutskever:2011generating}
Ilya Sutskever, James Martens, and Geoffrey~E Hinton. 2011.
\newblock Generating text with recurrent neural networks.
\newblock In \emph{Proceedings of the 28th International Conference on Machine
  Learning (ICML-11)}, pages 1017--1024.

\bibitem[{Vaswani et~al.(2017)Vaswani, Shazeer, Parmar, Uszkoreit, Jones,
  Gomez, Kaiser, and Polosukhin}]{vaswani:2017}
Ashish Vaswani, Noam Shazeer, Niki Parmar, Jakob Uszkoreit, Llion Jones,
  Aidan~N Gomez, {\L}ukasz Kaiser, and Illia Polosukhin. 2017.
\newblock Attention is all you need.
\newblock In \emph{Advances in Neural Information Processing Systems}, pages
  5998--6008.

\bibitem[{{Vinyals} et~al.(2012){Vinyals}, {Ravuri}, and
  {Povey}}]{vinyals2012:voicernns2}
Oriol {Vinyals}, Suman~V. {Ravuri}, and Daniel {Povey}. 2012.
\newblock \href {https://doi.org/10.1109/ICASSP.2012.6288816} {Revisiting
  recurrent neural networks for robust asr}.
\newblock In \emph{2012 IEEE International Conference on Acoustics, Speech and
  Signal Processing (ICASSP)}, pages 4085--4088.

\bibitem[{Xiong et~al.(2016)Xiong, Droppo, Huang, Seide, Seltzer, Stolcke, Yu,
  and Zweig}]{xiong2016achieving}
Wayne Xiong, Jasha Droppo, Xuedong Huang, Frank Seide, Mike Seltzer, Andreas
  Stolcke, Dong Yu, and Geoffrey Zweig. 2016.
\newblock Achieving human parity in conversational speech recognition.
\newblock \emph{arXiv preprint arXiv:1610.05256}.

\bibitem[{Yang et~al.(2017)Yang, Dai, Salakhutdinov, and Cohen}]{Zhilin:17}
Zhilin Yang, Zihang Dai, Ruslan Salakhutdinov, and William~W. Cohen. 2017.
\newblock \href {http://arxiv.org/abs/1711.03953} {Breaking the softmax
  bottleneck: {A} high-rank {RNN} language model}.
\newblock \emph{CoRR}, abs/1711.03953.

\bibitem[{Zhang and LeCun(2015)}]{zhang:2015text}
Xiang Zhang and Yann LeCun. 2015.
\newblock Text understanding from scratch.
\newblock \emph{arXiv preprint arXiv:1502.01710}.

\bibitem[{Zo{\l}na et~al.(2017)Zo{\l}na, Arpit, Suhubdy, and
  Bengio}]{zolna2017:fraternal}
Konrad Zo{\l}na, Devansh Arpit, Dendi Suhubdy, and Yoshua Bengio. 2017.
\newblock Fraternal dropout.
\newblock \emph{stat}, 1050:31.

\bibitem[{Zoph and Le(2016)}]{zoph2016:neural}
Barret Zoph and Quoc~V Le. 2016.
\newblock Neural architecture search with reinforcement learning.
\newblock \emph{arXiv preprint arXiv:1611.01578}.

\end{thebibliography}
\bibliographystyle{acl_natbib}

\end{document}